\documentclass[preprint,12pt]{elsarticle}

\usepackage{amssymb}
\usepackage{amsmath}
\usepackage{latexsym}
\usepackage{url}
\usepackage{xcolor}
\usepackage{epsfig}
\usepackage{graphicx}
\usepackage{bbding}
\usepackage{pifont}
\usepackage{wrapfig}
\usepackage{microtype}
\usepackage{amsfonts}
\usepackage{soul}
\usepackage{mathrsfs}
\usepackage{algorithmic,algorithm}
\usepackage{multirow}
\usepackage{subfigure}
\usepackage{array,caption}
\usepackage{bbm}
\usepackage{xcolor,colortbl}
\usepackage{booktabs}
\usepackage{makecell}
\usepackage{tikz}
\usepackage{xspace}
\usepackage[colorlinks, linkcolor=black, anchorcolor=black, citecolor=black]{hyperref}
\usepackage{adjustbox}
\usepackage{placeins}

\newcommand{\cmark}{\ding{51}}%
\newcommand{\xmark}{\ding{55}}%

\def\shortmodelname{Bro}

\journal{DLC 2026}

\begin{document}

\begin{frontmatter}




\title{Prototypical Few-Shot Medical Image Semantic Segmentation with Background Fusion}


\author[inst1]{Yuan Dong}
\author[inst2]{Xiaoyu Yu}
\author[inst1]{Wentao Wan}
\author[inst1]{Jianchao Xue}
\author[inst2]{Yuejin Duan}

\author[inst2]{Song Tang\corref{cor1}}
\ead{tangs@usst.edu.cn}

\author[inst2]{Yu Zhao\corref{cor1}}
\ead{zhaoyupumch@163.com}

\cortext[cor1]{Corresponding author}
\affiliation[inst1]{
    organization={Chinese Academy of Medical Sciences and Peking Union Medical College, Department of Orthopedics, Peking Union Medical College Hospital},
    city={Beijing},
    country={China}
}

\affiliation[inst2]{
    organization={School of Health Sciences and Engineering, University of Shanghai for Science and Technology},
    city={Shanghai},
    country={China}
}


\begin{abstract}
Few-shot Semantic Segmentation (FSS) aims to adapt a pre-trained model to new classes with as few as a single labeled training sample per class. 
The existing prototypical work used in natural image scenarios biasedly focus on capturing foreground's discrimination while employing a simplistic representation for background, grounded on the inherent observation separation between foreground and background. 
However, a frequency spectrum entropy analysis suggests that this paradigm is not applicable to medical images where the foreground and background share numerous visual features, necessitating a more detailed description for the background. 
In this paper, we present a new {{\bf B}ackground-fused p{\bf ro}totype} ({\bf \shortmodelname}) approach for FSS in medical images. 
Instead of identifying a commonality of background subjects in the support image, {\shortmodelname} fuses this background to discriminative prototypes, with two pivot designs. 
Specifically, Feature Similarity Calibration (FeaC) initially reduces noise in the support image by employing feature cross-attention with the query image. 
Subsequently, Hierarchical Channel-Adversarial Attention (HiCA) merges the background into comprehensive prototypes. 
We achieve this by a channel groups-based attention mechanism, where an adversarial Mean-Offset structure encourages a coarse-to-fine fusion. 
Designed as a generic plug-in, our {\shortmodelname} can be seamlessly integrated with existing FSS models. 
Extensive experiments validate the specificity of the background in medical images and the efficacy of {\shortmodelname} in enhancing the performance of previous FSS models on standard benchmarks.
\end{abstract}




\end{frontmatter}




\vspace{-5mm}
\section{Introduction}
\label{doc:intro}

Medical image segmentation is a foundational task in clinical processes and medical research, with significant potential for various downstream applications such as disease diagnosis~\citep{sherer2021metrics} and treatment planning~\citep{sherer2021metrics}. 
Among the current topics, Few-shot Semantic Segmentation~(FSS)~\citep{ouyang2022self} is an important area of focus, to account for the limited availability of well-annotated data, which arises from the protection of privacy and the requirement of clinical expertise. 
Unlike the conventional setting with segmentation labels, FSS's objective is to predict the tissue or organ in query data, the same as the given one or several support data.


\setlength{\parskip}{0pt}
In the view of building the similarity between the query and support images, the existing approaches mainly align to three lines: (1) The knowledge distillation framework~\citep{shaban2017one} (query and support images are inputs for the student and teacher branches, respectively), (2) the relevance structure discovering, e.g., attention~\citep{wang2020few} and graph~\citep{xie2021scale}, to identify shared features representing this similarity, and (3) the prototypical approach~\citep{wang2019panet} to generate prototypes from support images to build this similarity with the query image. 
Since the prototypes capture the discriminative and robust visual factors while being compatible with the classic convolution computation pipeline, the prototypical paradigm is widely applied. 
In practical design, the previous prototypical methods engage in extracting the discriminative foreground prototype, the same as the scenarios of natural images, while background is represented by simplistic schemes, such as Average Pooling~\citep{ouyang2022self} and feature filling~\citep{cheng2024few}. 
However, {\em Is foreground prototype sufficient for medical images?} 

\begin{figure}[!h]
    \centering
        \includegraphics[width=0.86\linewidth]{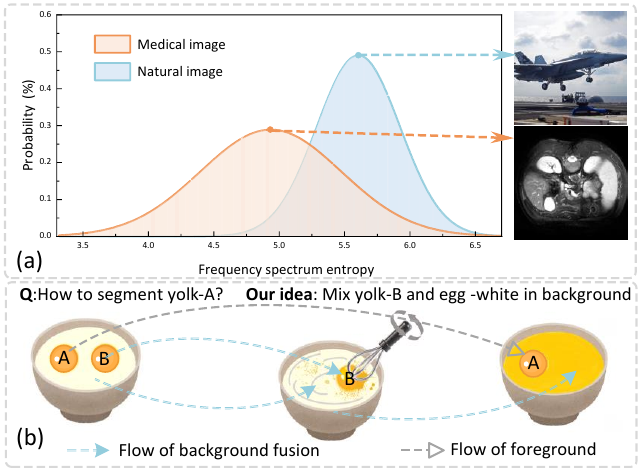}
    \caption{ 
    Motivation of {\shortmodelname}. (a) As shown in the comparison of probability distribution of frequency spectrum entropy (more experimental analysis is elaborated in \texttt{Sec.~\ref{sec:pdfse}}), the lower mean of medical images suggests a more concentrated frequency distribution than natural images. 
    Correspondingly, the background in medical images has more similar features to the foreground, necessitating a further background representation to discriminate it from the foreground.  
    To this end, we propose a background fusion scheme {\shortmodelname} whose idea is illustrated in (b) intuitively. 
    }
    \label{fig:ent-cmp}
\end{figure}

To clarify the issue above, we compare medical images with natural images based on the probability distribution of frequency spectrum entropy. 
As shown in Fig.~\ref{fig:ent-cmp}(a), notably, natural images have a significantly higher mean ({\bf 5.6}) compared to that of medical images. 
This higher mean indicates a flatter distribution across frequencies, suggesting that natural images maintain a balance between high-frequency foreground elements (the main subjects) and low-frequency backgrounds (see right top image). 
This inherent pattern allows for a clearer observation distinction between the foreground and background, justifying the foreground-centered approach commonly used in natural image scenarios. 
Conversely, medical images exhibit a lower mean ({\bf 4.9}), indicating that their frequency components are more concentrated. As a result, organs and tissues often share similar visual patterns.
For example, when considering the left kidney as the foreground target (see right bottom image), the background contains right kidney, gallbladder, spleen and in-between tissues with a similar texture as the foreground one, causing confusion in distinguishing between them.
{\em In short, the distinction between foreground and background is less pronounced in medical images than in natural images, necessitating a tailored design for background prototypes.}
In this paper, we introduce a new pluggable {\em {\bf B}ackground-fused p{\bf ro}totype} ({\bf \shortmodelname}) approach for FSS in medical images. 
One intuitive illustration of our idea is provided in Fig.~\ref{fig:ent-cmp}(b): Segmentation of yolk-A (foreground) in a two-yolk egg. 
Obviously, extracting the commonality of the egg-white with yolk-B (both in the background) presents a challenge.
Our solution is to thoroughly blend the components, allowing yolk-A to be smoothly separated from the resulting pale yellow mixture (the fused background).

In practice, we achieve the blending/fusion through joint usage of Feature-similarity Calibration (FeaC) module and Hierarchical Channel-adversarial Attention (HiCA) module.  
Specifically, FeaC first imposes a cross-attention between query and support feature maps, reducing the noise of low-similarity subjects in the support image.  
Following that, HiCA transforms the support image's background into fused prototypes by applying attention across channel groups. 
The attention mechanism constructs a similarity matrix for fusion in two steps: 
(1) It first identifies the coarse-grained similarity among those channel groups using self-cross-similarity calculation, and then (2) it fine-tunes this similarity utilizing an adversarial strategy building on a Mean-Offset structure.


Our {\bf contributions} are summarized as: 
\begin{itemize}

\item We examine the necessity of background representation for prototypical FSS in medical images and propose a new representation scheme of fusing background in the support image, which conceptually differs from the previous strategy of extracting cross-subject commonality. 



\item We propose a novel adversarial attention approach {\shortmodelname} to achieve this background fusion. {\shortmodelname} is a generic plug-in module, which is featured by a channel group-based attention, building upon an adversarial Mean-Offset structure. 




\item We integrate {\shortmodelname} with the previous state-of-the-art methods and perform extensive evaluations on three medical benchmarks. The evident performance gains compared to the original approaches validate the effectiveness of {\shortmodelname}.
\end{itemize}



\vspace{-3mm}
\section{Related Work}
\paragraph{Medical image segmentation}
Currently, the landscape of medical image segmentation is heavily dominated by deep neural network paradigms. The inception of this field borrowed extensively from natural image semantic segmentation techniques. A seminal milestone was achieved by Fully Convolutional Networks (FCNs)~\citep{long2015fully}, which adapted standard Convolutional Networks (CNNs) for pixel-level prediction by incorporating up-sampling mechanisms and skip layers. To address the coarse boundary reconstructions inherent in FCNs, subsequent researchers pioneered symmetrical encoder-decoder frameworks, such as SegNet~\citep{badrinarayanan2017segnet} and DeconvNet~\citep{noh2015learning}, which excel at capturing intricate semantic details. As deep learning matured within the clinical domain, specialized medical imaging architectures began to surface. Chief among these is U-Net~\citep{ronneberger2015u}, which has garnered widespread acclaim for its exceptional efficacy. Beyond its balanced encoder-decoder topology, U-Net uniquely integrates skip connections to seamless transfer high-resolution contextual cues across feature hierarchies. This groundbreaking paradigm has catalyzed a myriad of sophisticated derivatives, notably 3D U-Net~\citep{cciccek20163d}, Attention U-Net~\citep{oktay2018attention}, Edge-U-Net~\citep{allah2023edge}, V-Net~\citep{milletari2016v}, and Y-Net~\citep{mehta2018net}.


These segmentation models above only work in a supervised fashion, relying on abundant expert-annotated data. Thus, they cannot apply to the few-shot setting where we need to segment an object of an ``unseen" class as only a few labeled images of this class are given.


\vspace{0.1cm}
\noindent{\bf Few-shot semantic segmentation.}
The existing FSS methods follow three lines according to the idea of building a class-wise similarity between the query and support images. The first constructs a teacher-student branches-based framework~\citep{shaban2017one,roy2020squeeze,sun2022few}, where the support images-based guidance (teacher)
is used to regulate segmentation branch over the query image (student). The second designed novel network modules, e.g., attention modules~\citep{hu2019attention,wang2020few}, graph networks~\citep{gao2022mutually,xie2021scale}, and representative descriptors~\citep{cheng2024few}, for discriminative representations,by which the features shared by query and support images were identified. The third is the prototypical line, which constructs foreground/background prototypes to bridge the similarity computation in a meta-learning fashion, such as dual-directive prototype alignment~\citep{wang2019panet}, region-enhanced prototypical transformers~\citep{zhu2023rpt}, de-biased prototypes~\citep{zhu2024learning}, prototype correlation matching with class-relation reasoning~\citep{zhang2024pmcr}, and high-fidelity prototype learning~\citep{Tang2024FewShotMI}. 
Recently, self-supervised and training-efficient FSMIS methods have been further explored, including superpixel-based self-supervision~\citep{ouyang2020self,ouyang2022self}, anomaly-inspired pseudo-labeling~\citep{hansen2022anomaly}, hard-prototype mining for boundary-sensitive regions~\citep{jiang2025cow}, and large-kernel attention for support-query feature enhancement~\citep{wu2024largekernel}. Meanwhile, training-free foundation-model-based methods, such as SAM2-based support-query matching~\citep{zu2025sam2}, have also shown promising performance.

These works improve the utilization of limited annotations from different perspectives, but most of them still focus on foreground prototypes or support-query matching, leaving the comprehensive modeling of complex medical backgrounds less explored.


\setlength{\parskip}{0pt}
The proposed model, {\shortmodelname}, is a prototypical approach that aligns with the third line and differs from previous work in two key ways.
First, {\shortmodelname} emphasizes the importance of representing the background in medical images, an aspect that earlier methods often overlooked. 
Second, rather than extracting common features, {\shortmodelname} employs a fusion strategy to create a comprehensive representation. 


\vspace{-4mm}
\section{Methodology}\label{sec:method}

\subsection{Problem Formulation of FSS}

The FSS setting involves two datasets without shared categories: The training subset $\mathcal{D}_{tr}$ (annotated by $\mathcal{Y}_{tr}$) and the test subset $\mathcal{D}_{te}$ (annotated by $\mathcal{Y}_{te}$), both of which consist of image-mask pairs and $\mathcal{Y}_{tr}\cap \mathcal{Y}_{te}= \varnothing$.  
The goal of FSS is to train a segmentation model on $\mathcal{D}_{tr}$ that can segment unseen semantic classes $\mathcal{Y}_{te}$ in images in $\mathcal{D}_{te}$, given a few annotated examples of $\mathcal{Y}_{te}$, without re-training.


This paper approaches the FSS problem using a meta-learning framework, similar to initial few-shot segmentation methods. 
We sclice $\mathcal{D}_{tr}= \{S_{i},Q_{i}\}_{i=1}^{N_{tr}}$ into several randomly sampled episodes, as well as $\mathcal{D}_{te}= \{S_{i},Q_{i}\}_{i=1}^{N_{te}}$. Here, ${N_{tr}}$ and ${N_{te}}$ are the episode numbers for training and testing, respectively.
Thus, each episode consists of $K$ annotated support images and a collection of query images containing $N$ categories.
Specifically, we define this as an \(N\)-way \(K\)-shot segmentation sub-problem. 
The support set \(S_{i} = \{({I}_{k}^{s}, {m}_{k}^{s}({c}_{j}))\}_{k=1}^{K}\) includes \(K\) image-mask pairs, where \({I}_{k}^{s}\) is a grayscale image in \(\mathbb{R}^{H \times W}\) and its corresponding binary mask \({m}_{k}^{s} \in \{0, 1\}^{H \times W}\) is for class \(c_{j} \in C_{tr}\), with \(j=1, 2, \ldots, N\).
The query set ${Q_{i}}$ contains ${V}$ image-mask pairs from the same class as the support set. 
While the training on $\mathcal{D}_{tr}$, over each episode, we learn a function $f(I^q, S_i)$, which predicts a binary mask of an unseen class when given the query image $I^q \in {Q_{i}}$ and the support set $S_i$.  
After completing a series of episodes, we obtain the final segmentation model, which is evaluated on ${N_{te}}$ in the same $N$-way $K$-shot segmentation manner. 
Following the common practice in~\citep{ouyang2020self, achanta2012slic, shen2023qnet}, this paper set $N=K=1$.

\begin{figure*}[!h]
    \setlength{\belowcaptionskip}{-0.3cm}
    \setlength{\abovecaptionskip}{0pt}
    \begin{center}
        \includegraphics[width=1\linewidth]{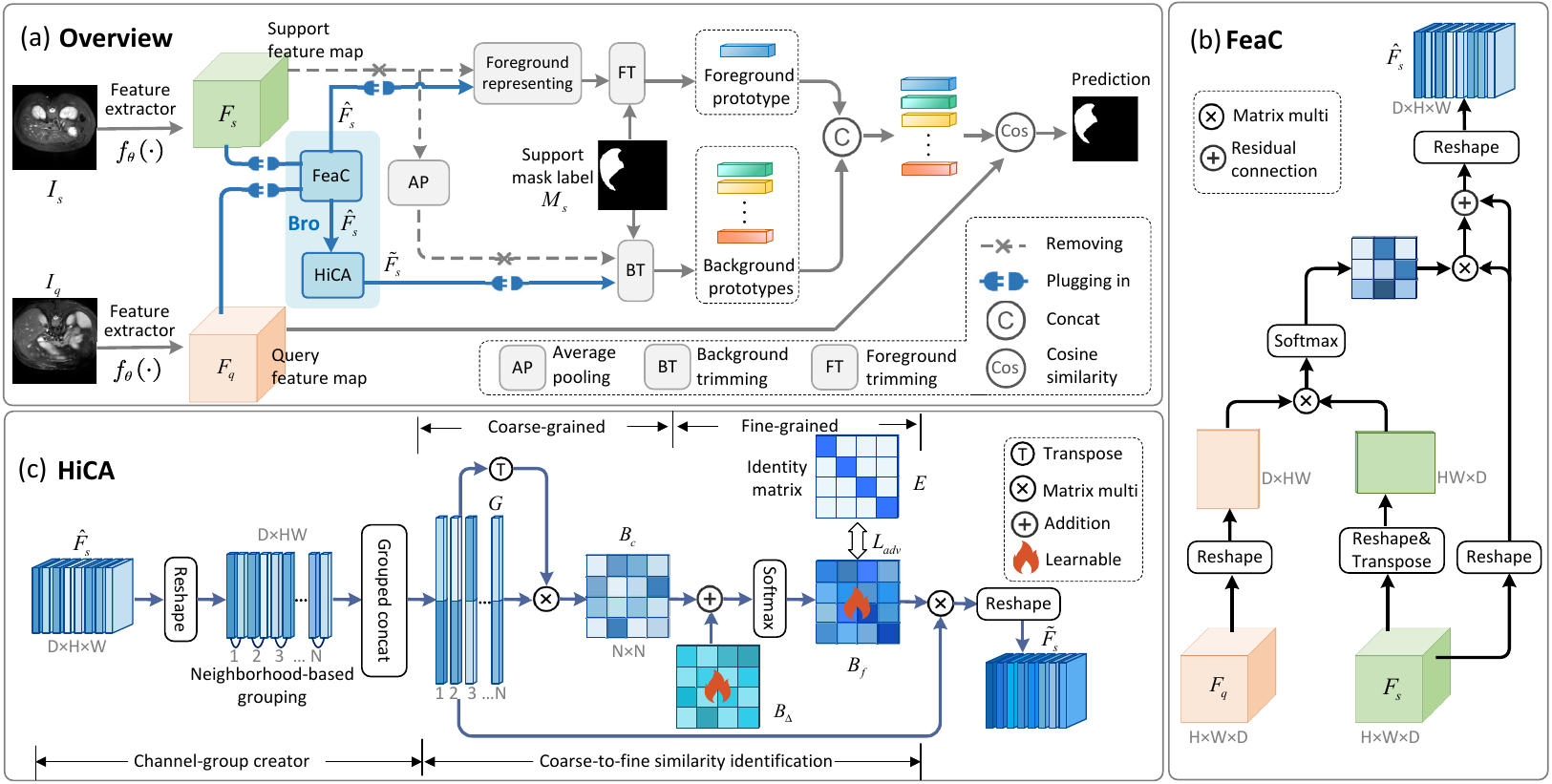}
    \end{center}
    \caption{
    Overview of the SSL-ALPNet framework plugged with {\shortmodelname}. (a) Unlike directly trimming background prototypes in the conventional pipeline (marked by gray lines), {\shortmodelname} provides an ability of discriminative background representation. 
    In this module, (b) FeaC denoise support feature map ${F}_s$ by calibrating similarity with query feature map ${F}_q$. After that, (c) HiCA generates detailed background representation $\widetilde{F}_S$ by performing a channel group attention-based fusion over the similarity calibrated $\hat{F}_S$.}
    \label{fig:fw}
\end{figure*}

\subsection{Overview}
In this paper, without loss of generality, we present {\shortmodelname} based on the famous self-supervised framework SSL-ALPNet~\citep{ouyang2022self}. 
As depicted in Fig.~\ref{fig:fw} (a), the segmentation pipeline follows (1) Feature extraction $f(\cdot)$, (2) prototypes generation plugged with {\shortmodelname} consisting of the FeaC and HiCA modules, and (3) segmentation with cosine similarity.

Specifically, suppose that the support and query images are denoted by $I_s$ and $I_q$, respectively. 
The segmentation begins with feature extraction $F_s=f_{\theta}(I_s)$ and $F_q=f_{\theta}(I_q)$ ($\theta$ is the model parameters), followed by the similarity calibration module FeaC, which fuses $F_s$ and $F_q$ to $\hat{F}_s$. 
Subsequently, the generation of foreground prototype $P_f$ is the same as SSL-ALPNet. 
That is, the foreground representation is achieved by the adaptive local representation method, while the foreground trimming is implemented using Masking Average Pooling with support masking label $M_s$.
At the same time, HiCA produces a background-fused feature map $\widetilde{F}_s$, replacing the previous background representation based on the Average Pooling operation.   
The background trimming tailors $P_b$ from $\widetilde{F}_s$ using the background zone in $M_s$, the same as SSL-ALPNet.   
Finally, we obtain the query prediction of segmentation by calculating cosine similarity between $F_q$ and generated prototypes $\{P_f, P_b\}$ in a convolutional way. 
The details of FeaC and HiCA are presented below. 

{\bf Remark:} In the proposed framework shown in Fig.~\ref{fig:fw}, the foreground block is not restricted to the strategy in SSL-ALPNet; alternative approaches may also be utilized.

\subsection{Feature Similarity Calibration} 
In the FeaC module, we calibrate the similarity between $F_s$ and $F_q$ employing a cross-attention structure with a residual connection, as shown in Fig.~\ref{fig:fw} (b), which is widely used in feature integration~\citep{wang2018non,huang2019ccnet,ding2023few}.
Formally, suppose that the support and query feature maps $F_s$, $F_q$ are reshaped to matrix ${U}_s$ and ${U}_q$, respectively. This attention mechanism can be formulated as:
\begin{equation}
    \label{eqn:feac}
    \begin{split} 
    \hat{{F}}_{s} = \frac{{{\delta}\left({U}_{s}^{T}\times {U}_{q}\right)}\times{U}_{s}}{\left \| {U}_{s}\right \|\left \|{U}_{q} \right \|}+ {U}_{s},  
  \end{split}
\end{equation}
where $\delta(\cdot)$ stands for softmax operation, $\times$ means matrix multiplication, ${\delta }\left({U}_{s}^{T}\times {U}_{q}\right)$ means the similarity-based probability matrix weighting $U_s$.

In the segmentation pipeline in Fig.~\ref{fig:fw} (a), FeaC serves as a precursor module that facilitates the generation of successive prototypes.
First, it enhances similar regions between $F_s$ and $F_q$, improving the foreground in $F_s$ that exists within those areas. 
Most importantly, it reduces irrelevant textures and objects between $F_s$ and $F_q$ helps to filter out background noise in $F_s$.

\subsection{Hierarchical Channel-adversarial Attention}
As mentioned earlier, a challenge of FSS in medical images is representing background tissues and objects that are often confused with the foreground due to similar textures or shapes. 
Unlike natural categories, such as pig and cat, the term ``background" is an artificial, task-specific concept that encompasses multiple categories. 
Furthermore, each image typically features a unique background that corresponds to distinct categories. 
Therefore, it is questionable whether we can identify commonalities among these background subjects, the same as natural category representations.
In light of this, we propose the background fusion representation scheme HiCA. 

\vspace{0.1cm}
\noindent{\bf Overview.}
In our segmentation pipeline, HiCA executes the background fusion.
Its working process is detailed in Fig.~\ref{fig:fw} (c). 
For the input similarity calibrated feature map $\hat{F}_s$, the {\em channel-group creator} module produces channel groups $G$ from $\hat{F}_s$, denoted by $G={\rm{CG}}(\hat{F}_s)$. 
Following this, the {\em coarse-to-fine similarity identification} module generates a similarity matrix $B_f=g(G)$, combining the self-cross similarity calculation and adversarial regularization. 
Ultimately, we complete the fusion based on $B_f$, obtaining background-fused feature map $\widetilde{F}_s$. 
The proposed attention mechanism can be formulated as 
\vspace{-1mm}
\begin{equation}\label{eqn:hica-ts}
    \begin{split} 
    \widetilde{F}_{s} = g(G) \times {G},~~
    G={\rm{CG}}(\hat{F}_s). 
  \end{split}
\end{equation}
\vspace{-5mm}

As presented in Eq.~\eqref{eqn:hica-ts}, HiCA achieves fusion based on channel dimension attention rather than feature dimension approach, e.g., FeaC. 
The rationale behind this design is as follows.
Similar to the frequency domain, the visual factors associated with these channels are often located across different subjects. 
Within this context, channel attention identifies the channels that are positively relevant to segmentation, equivalently leading to a dense sampling across those subjects located over the entire image. 
In other words, it facilitates a specific fusion of information within the image.
In contrast, feature attention leads to convergence to some isolated regions within spatial dimensions.


\vspace{0.1cm}
\noindent{\bf Channel group creator.}
As shown in the left side of Fig.~\ref{fig:fw} (c), we obtain channel groups $G$ according to the following workflow.  
Suppose $\hat{F}_s\in \mathbb{R}^{D \times W\times H}$ is reshaped to channel vectors $\{\boldsymbol{c}_i\in \mathbb{R}^{1 \times W*H}\}_{i=1}^{D}$, and further converted to $\{\boldsymbol{g}_i\in \mathbb{R}^{1 \times W*H*N}\}_{i=1}^{D/N}$ by grouping neighboring $N$ channel vectors without overlap and concatenating them together. 
Formally, $\boldsymbol{g}_i = {\rm{concat}}\left(\{{\boldsymbol{c}}_{(i-1)*N+1}, \cdots, {\boldsymbol{c}}_{i*N}\}\right)$. 
Collectively writing $\{\boldsymbol{g}_i\}_{i=1}^{D/N}$ to matrix form, we obtain $G$.


\vspace{0.1cm}
\noindent{\bf Coarse-to-fine similarity identification.}
The generation of similarity matrix $B_f$ involves two steps: (1) First obtaining the coarse-grained similarity matrix $B_c$ by self-cross correlation calculation and then (2) adjusting $B_c$ in an adversarial manner, resulting in the final fine-grained similarity. 
We propose a Mean-Offset structure to achieve this goal, which is summarized to the following equation. 
\begin{equation}\label{eqn:hica} 
    \begin{split} 
    B_f = g(G) = \frac{\delta(\overbrace{{G}^{T}\times {G}}^{B_c} + \alpha B_{\triangle})}{{\left \| {G}^T\right \|\left \|{G} \right \|\left \|{B_{\triangle}} \right \|}},
  \end{split}
\end{equation}
where $\delta(\cdot)$ and $\times$ means softmax operation and matrix multiplication, respectively; $B_{\triangle}$ is similarity offset matrix, parameter $\alpha$ is adjustment strength.

 

In Eq.~\eqref{eqn:hica}, by combining matrix $B_c$ with the trainable offset $B_{\triangle}$, we realize the coarse-to-fine identification of similarities. However, the unrestricted adjustments to \(B_{\triangle}\) might disrupt the representational relationship between channels and their associated semantics. To address this issue, we propose an adversarial regulation formulated as:   
\begin{equation}\label{eqn:adver}
\small
    \begin{split} 
    \mathcal{L}_{adv} = \left \| B_f - E \right \|_2= \sum_i (B_f^{ii}-1)^2 + \sum_i \sum_{i\neq k} (B_f^{ik})^2,
  \end{split}
\end{equation}
where $E$ is the Identity matrix, $B_f^{ii}$ and $B_f^{ik}$  are the diagonal and non-diagonal elements in $B_f$, respectively. 
Here, minimizing $\mathcal{L}_{ad}$ encourages these channel groups to be independent of each other, making the channel groups converge to the original representational relationship.
Thus, this regularization provides a reverse optimization direction (enforcing off-diagonal elements are 0) to depress the adjustment of $B_{\triangle}$. 
Within this adversarial context, we can capture an optimal similarity between the channel groups.


{\bf Remark.} The HiCA module differentiates itself from earlier channel-attention methods in two key design aspects. First, it focuses attention on groups of channels instead of just one channel at a time. Second, it includes adversarial regularization. These features are inspired by the link between frequency and visual elements. For instance, the high-frequency band is typically associated with foreground details. By treating each channel group as a frequency band, we can assign specific semantics to each group. This approach allows for improved semantic fusion by focusing attention on these channel groups. More importantly, adjusting or maintaining this representational relationship fosters an adversarial balance.

\subsection{Training Objective}
FSS is a pixel-level classification task, thereby adopting cross-entropy loss to regulate model training. 
\begin{equation} \label{eqn:loss_seg}
    \small
    \begin{aligned}
    \mathcal{L}_{seg} \!=\!  
    - \frac{1}{HW}\displaystyle\sum_{h}^{H}\displaystyle\sum_{w}^{W}
    \displaystyle\sum_{j\in \{f,b\}}^{} {m}_{q}^{j}\left(h,w\right) \odot log\left(\hat{m}_{q}^{j}\left(h,w\right)\right),
    \end{aligned}
\end{equation}
where $\hat{m}_{q}^{j}(h,w)$ is the predicted results of the query mask label ${m}_{q}^{j}(h,w)$; in $\{f,b\}$, $f$ and $b$ means foreground and background, respectively. 
In addition, the same as~\citep{wang2019panet,ouyang2022self,shen2023qnet}, we regulate another inverse learning encouraging a prototypical alignment. 
In practice, the query images serve as the support set to predict labels of the support images. 
The alignment regularization is expressed as 
\begin{equation} \label{eqn:loss_reg} 
    \small
    \begin{aligned}
    \mathcal{L}_{reg} =  
    - \frac{1}{HW}\displaystyle\sum_{h}^{H}\displaystyle\sum_{w}^{W}
    \displaystyle\sum_{j\in \{f,b\}}^{} {m}_{s}^{j}(h,w) \odot log\left(\hat{m}_{s}^{j}(h,w)\right).
    \end{aligned}
\end{equation}
Finally, combining the adversarial loss in Eq.~\eqref{eqn:adver}, the model training is summarized to the optimization problem below.
\begin{equation}\label{eqn:loss_final}
    \begin{split} 
    \min_{\{\theta, B_{\triangle}, B_{f}\}}^{} \mathcal{L}_{seg} + \mathcal{L}_{reg} + \beta \mathcal{L}_{adv}. 
    \end{split}
\end{equation}
where $\beta$ is a trade-off parameter, $\theta$ is the feature extractor parameters. 
Due to following the self-supervision fashion, we do not provide the real masks of query and support images ($m_q$ in Eq.~\eqref{eqn:loss_seg} and $m_s$ in Eq.~\eqref{eqn:loss_reg}). Instead, we generated pseudo masks by Superpixels method, as same to \cite{ouyang2022self}.

\FloatBarrier


\vspace{-2mm}
\section{Experiments}

\subsection{Data Sets}

To demonstrate the effectiveness of the proposed method, we conduct evaluation on three challenging medical benchmarks. 
Their details are presented as follows.


\begin{itemize}
\item  \textbf{Abdominal CT dataset}~\citep{r31}, termed {\bf ABD-CT}, was acquired from the Multi-Atlas Abdomen Labeling challenge at the Medical Image Computing and Computer Assisted Intervention Society (MICCAI) in 2015. This dataset contains 30 3D abdominal CT scans. Of note, this is a clinical dataset containing patients with various pathology's and variations in intensity distributions between scans.

\item \textbf{Abdominal MRI dataset}~\citep{r32}, termed {\bf ABD-MRI}, was obtained from the Combined Healthy Abdominal Organ Segmentation (CHAOS) challenge held at the IEEE International Symposium on Biomedical Imaging (IS BI) in 2019. This dataset consists of 20 3D MRI scans with a total of four different labels representing different abdominal organs.

\item  \textbf{Cardiac MRI dataset}~\citep{r33}, termed {\bf CMR}, was obtained from the Automatic Cardiac Chamber and Myocardium Segmentation Challenge held at the Conference on Medical Image Computing and Computer Assisted Intervention (MICCAI) in 2019. It contains 35 clinical 3D cardiac MRI scans.
\end{itemize}

\begin{table*}[!t]
    \caption{Results on the {\bf ABD-MRI} and {\bf ABD-CT} datasets.  
    Numbers in bold indicated the best results.}
    \label{tab:rlt-ABD}
    \renewcommand\tabcolsep{6.5pt} 
    \renewcommand\arraystretch{1.2} 
    \centering
    \scriptsize
    \begin{adjustbox}{max width=\textwidth}
        \begin{tabular}{c|l|c|c|c|c|c|c|c|c|c|c}
        \toprule
        \multirow{2}{*}{\textbf{Setting}} &
        \multirow{2}{*}{\textbf{Method}} &
        \multicolumn{5}{c|}{\textbf{ABD-MRI}} &
        \multicolumn{5}{c}{\textbf{ABD-CT}}\\

        \cline{3-12}

        &
        & \textbf{Liver}
        & \textbf{R.kidney}
        & \textbf{L.kidney}
        & \textbf{Spleen}
        & \textbf{Mean}
        & \textbf{Liver}
        & \textbf{R.kidney}
        & \textbf{L.kidney}
        & \textbf{Spleen}
        & \textbf{Mean} \\

        \hline

        \multirow[c]{12}{*}{\makecell{Setting-1}}

        & PANet~\citep{wang2019panet}
        &47.37&30.41&34.96&27.73&35.11
        &60.86&50.42&56.52&55.72&57.88\\

        & ADNet~\citep{hansen2022anomaly}
        &76.79&84.21&62.97&49.74&68.42
        &77.26&32.86&31.27&41.17&45.67\\

        & RPTNet~\citep{zhu2023rpt}
        &60.32&86.83&65.58&73.72&71.61
        &64.51&60.52&84.37&68.48&69.47\\

        & Q-Net~\citep{shen2023qnet}
        &72.47&86.40&72.13&76.24&76.81
        &68.65&55.63&69.39&56.82&62.63\\

        & CAT-Net~\citep{lin2023few}
        &70.59&83.00&75.30&70.54&74.86
        &66.24&47.83&69.09&66.98&62.54\\

        & GMRD~\citep{cheng2024few}
        &73.65&89.95&75.97&65.44&76.25
        &63.06&62.27&79.92&56.48&65.43\\

        \cline{2-12}

        & SSL-ALPNet~\citep{ouyang2022self}
        &74.32&84.88&79.61&67.78&76.65
        &67.29&72.62&76.35&70.11&71.59\\

        &\cellcolor{gray!40}\textbf{SSL-ALPNet+\shortmodelname}
        &\cellcolor{gray!40}74.30
        &\cellcolor{gray!40}87.06
        &\cellcolor{gray!40}83.49
        &\cellcolor{gray!40}68.13
        &\cellcolor{gray!40}78.25
        (\textcolor{red}{\textbf{$+$1.60}})
        &\cellcolor{gray!40}71.01
        &\cellcolor{gray!40}79.07
        &\cellcolor{gray!40}77.91
        &\cellcolor{gray!40}71.71
        &\cellcolor{gray!40}74.93
        (\textcolor{red}{\textbf{$+$3.34}})\\
        & DSPNet~\citep{Tang2024FewShotMI}
        &{73.85}  &{84.40}  &{83.38}  &{61.52}  &{75.79}
        &{69.32}  &{74.54}  &{78.01}  &{69.31}  &{72.79}\\

        &\cellcolor{gray!40}\textbf{DSPNet+\shortmodelname}
        &\cellcolor{gray!40}{70.64}
        &\cellcolor{gray!40}{85.11}
        &\cellcolor{gray!40}{80.53}
        &\cellcolor{gray!40}{72.41}
        &\cellcolor{gray!40}{77.17}
        (\textcolor{red}{\textbf{$+$1.38}})
        &\cellcolor{gray!40}{69.00}
        &\cellcolor{gray!40}{77.06}
        &\cellcolor{gray!40}{79.04}
        &\cellcolor{gray!40}{67.88}
        &\cellcolor{gray!40}{73.24}
        (\textcolor{red}{\textbf{$+$0.45}})
        \\

        & COW~\citep{jiang2025cow}
        &\textcolor{blue}{\textbf{81.52}}
        &{89.06}
        &{81.19}
        &{75.41}
        &{81.80}
        &{78.78}
        &{80.66}
        &{82.70}
        &\textcolor{blue}{\textbf{82.02}}
        &{81.04}\\

        &\cellcolor{gray!40}\textbf{COW+\shortmodelname}
        &\cellcolor{gray!40}{79.63}
        &\cellcolor{gray!40}\textcolor{blue}{\textbf{90.88}}
        &\cellcolor{gray!40}\textcolor{blue}{\textbf{84.43}}
        &\cellcolor{gray!40}\textcolor{blue}{\textbf{79.17}}
        &\cellcolor{gray!40}\textcolor{blue}{\textbf{83.53}}
        (\textcolor{red}{\textbf{$+$1.73}})
        &\cellcolor{gray!40}\textcolor{blue}{\textbf{80.49}}
        &\cellcolor{gray!40}\textcolor{blue}{\textbf{81.96}}
        &\cellcolor{gray!40}\textcolor{blue}{\textbf{85.08}}
        &\cellcolor{gray!40}{81.11}
        &\cellcolor{gray!40}\textcolor{blue}{\textbf{82.16}}
        (\textcolor{red}{\textbf{$+$1.12}})
        \\

        \hline
        \hline

        \multirow{12}{*}{\makecell{Setting-2}}

        & PANet~\citep{wang2019panet}
        &{69.37}  &{66.94}  &{63.17}  &{61.25}  &{65.68}
        &{61.71}  &{34.69}  &{37.58}  &{43.73}  &{44.42}\\

        & ADNet~\citep{hansen2022anomaly}
        &{77.03}  &{59.64}  &{56.68}  &{59.44}  &{63.19}
        &{70.63}  &{48.41}  &{40.52}  &{50.97}  &{52.63}\\

        & RPTNet~\citep{zhu2023rpt}
        &{67.45}  &{60.11}  &{66.27}
        &\textcolor{blue}{\textbf{75.15}}
        &{67.25}
        &{54.24}
        &{53.84}
        &{82.28}
        &{60.12}
        &{62.62}\\

        & Q-Net~\citep{shen2023qnet}
        &{71.52}
        &{74.71}
        &{64.15}
        &{74.71}
        &{71.27}
        &{64.44}
        &{41.75}
        &{66.21}
        &{37.87}
        &{52.57}\\

        & CAT-Net~\citep{lin2023few}
        &{77.45}
        &{60.23}
        &{78.57}
        &{60.23}
        &{69.12}
        &{52.53}
        &{46.87}
        &{65.01}
        &{46.73}
        &{52.79}\\

        & GMRD~\citep{cheng2024few}
        &{74.85}
        &{70.25}
        &{69.37}
        &{73.80}
        &{72.07}
        &{60.88}
        &{55.35}
        &{72.46}
        &{64.16}
        &{63.21}\\

        \cline{2-12}

        & SSL-ALPNet~\citep{ouyang2022self}
        &{68.38}
        &{76.38}
        &{73.24}
        &{55.35}
        &{68.34}
        &{69.14}
        &{59.05}
        &{64.18}
        &{61.97}
        &{63.59}\\

        &\cellcolor{gray!40}\textbf{SSL-ALPNet+\shortmodelname}
        &\cellcolor{gray!40}{69.55}
        &\cellcolor{gray!40}{81.94}
        &\cellcolor{gray!40}\textcolor{blue}{\textbf{81.40}}
        &\cellcolor{gray!40}{61.31}
        &\cellcolor{gray!40}{73.55}
                (\textcolor{red}{\textbf{$+$5.21}})
        &\cellcolor{gray!40}{60.81}
        &\cellcolor{gray!40}{66.81}
        &\cellcolor{gray!40}{65.49}
        &\cellcolor{gray!40}{67.87}
        &\cellcolor{gray!40}{65.24}
                (\textcolor{red}{\textbf{$+$1.65}})\\
        & DSPNet~\citep{Tang2024FewShotMI}
        &{71.46}
        &{75.87}
        &{78.49}
        &{66.00}
        &{72.96}
        &{69.16}
        &{63.55}
        &{68.46}
        &{66.48}
        &{66.17}\\

        &\cellcolor{gray!40}\textbf{DSPNet+\shortmodelname}
        &\cellcolor{gray!40}{75.41}
        &\cellcolor{gray!40}{80.00}
        &\cellcolor{gray!40}{77.69}
        &\cellcolor{gray!40}{69.06}
        &\cellcolor{gray!40}{75.54}
                (\textcolor{red}{\textbf{$+$2.58}})
        &\cellcolor{gray!40}{72.19}
        &\cellcolor{gray!40}{69.16}
        &\cellcolor{gray!40}{70.16}
        &\cellcolor{gray!40}{61.83}
        &\cellcolor{gray!40}{68.34}
                (\textcolor{red}{\textbf{$+$2.17}})\\

        & COW~\citep{jiang2025cow}
        &\textcolor{blue}{\textbf{81.51}}
        &\textcolor{blue}{\textbf{90.48}}
        &{75.44}
        &{73.60}
        &{80.26}
        &{81.35}
        &{77.69}
        &\textcolor{blue}{\textbf{84.07}}
        &{80.46}
        &{80.89}\\

        &\cellcolor{gray!40}\textbf{COW+\shortmodelname}
        &\cellcolor{gray!40}{81.40}
        &\cellcolor{gray!40}{90.14}
        &\cellcolor{gray!40}{79.50}
        &\cellcolor{gray!40}{74.19}
        &\cellcolor{gray!40}\textcolor{blue}{\textbf{81.31}}
                (\textcolor{red}{\textbf{$+$1.05}})
        &\cellcolor{gray!40}\textcolor{blue}{\textbf{82.02}}
        &\cellcolor{gray!40}\textcolor{blue}{\textbf{81.59}}
        &\cellcolor{gray!40}{83.53}
        &\cellcolor{gray!40}\textcolor{blue}{\textbf{80.52}}
        &\cellcolor{gray!40}\textcolor{blue}{\textbf{81.91}}
                (\textcolor{red}{\textbf{$+$1.02}})\\

        \bottomrule
    \end{tabular}

    \end{adjustbox}
\end{table*}

\begin{table}[!t]
    \caption{Results on {\bf CMR}. 
    Numbers in bold indicated the best results.}
    \label{tab:rlt-CMR} 
    \renewcommand\tabcolsep{22pt} 
    \renewcommand\arraystretch{1.2}
    \footnotesize
    \centering 
    \begin{adjustbox}{max width=\linewidth}
    \begin{tabular}{c|l|c|c|c|c}
    \toprule
    \textbf{Settings} & \textbf{Method} & \textbf{RV} & \textbf{LV-MYO} & \textbf{LV-BP} & \textbf{Mean}\\
    \hline
    
    \multirow[c]{12}{*}{\makecell[c]{Setting-1}}
    
    & PANet~\citep{wang2019panet}
    &57.13&44.76&72.77&58.20\\
    
    & ADNet~\citep{hansen2022anomaly}
    &65.37&\textcolor{blue}{\textbf{82.29}}&58.86&68.84\\
    
    & RPTNet~\citep{zhu2023rpt}
    &76.63&80.15&58.81&71.86\\
    
    & Q-Net~\citep{shen2023qnet}
    &67.99&52.09&86.21&68.76\\
    
    & CAT-Net~\citep{lin2023few}
    &69.37&48.81&81.33&66.51\\
    
    & GMRD~\citep{cheng2024few}
    &80.82&73.65&60.83&71.77\\
    
    \cline{2-6}
    
    & SSL-ALPNet~\citep{ouyang2022self}
    &77.59&63.29&85.36&75.41\\
    
    &\cellcolor{gray!20}\textbf{SSL-ALPNet+\shortmodelname}
    &\cellcolor{gray!20}78.39
    &\cellcolor{gray!20}63.29
    &\cellcolor{gray!20}87.79
    &\cellcolor{gray!20}76.49
            (\textcolor{red}{\textbf{$+$1.08}})

    \\
    
    & DSP-Net~\citep{Tang2024FewShotMI}
    &79.73&64.91&87.75&77.46\\
    
    &\cellcolor{gray!20}\textbf{DSP-Net+\shortmodelname}
    &\cellcolor{gray!20}80.30
    &\cellcolor{gray!20}65.96
    &\cellcolor{gray!20}88.86
    &\cellcolor{gray!20}78.37
            (\textcolor{red}{\textbf{$+$0.91}})

    \\
    
    & COW~\citep{jiang2025cow}
    &79.25&67.10&\textcolor{blue}{\textbf{89.09}}&78.48\\
    
    &\cellcolor{gray!20}\textbf{COW+\shortmodelname}
    &\cellcolor{gray!20}\textcolor{blue}{\textbf{82.10}}
    &\cellcolor{gray!20}67.59
    &\cellcolor{gray!20}88.40
    &\cellcolor{gray!20}\textcolor{blue}{\textbf{79.54}}
            (\textcolor{red}{\textbf{$+$1.06}})
    \\
    
    \bottomrule
    \end{tabular}

    \end{adjustbox}
\end{table}

\subsection{Implementation Details}

\paragraph{Few-shot setting} 
We follow the experimental settings in~\citep{ouyang2022self,hansen2022anomaly}, considering two cases. {\bf Setting-1} is the initial setting proposed in~\citep{roy2020squeeze}, where test classes may appear in the background of training images. We train and test on all classes in the dataset without any partitioning. {\bf Setting-2} is a strict version of Setting-1, proposed in~\citep{ouyang2022self}, where we adopted a stricter approach. In this setting, test classes do not appear in any training images. For instance, when segmenting Liver during training, the support and query images do not contain the Spleen, which is the segmenting target for testing. We directly removed the images containing test classes during the training phase to ensure that the test classes are truly ``unseen” for the model.

\paragraph{Network backbone \& pseudo masking label}
In all experiments, a fully convolutional Resnet101 model is taken as the feature extractor, being pre-trained on the MS-COCO dataset.
Given that the superpixel pseudo-labels contain rich clustering information, which are helpful to alleviate the annotation absence.
We generate the superpixel pseudo-label in an offline manner as the support image mask before starting the model training, following~\citep{ouyang2020self,shen2023qnet}.

\paragraph{Hyper-parameter setting}
The proposed {\shortmodelname} involves three parameters: $\alpha$ in Eq.~\eqref{eqn:hica}, $\beta$ in Eq.~\eqref{eqn:loss_final} and the grouping parameter $N$. 
In the {ABD-MRI} and {ABD-CT} datasets, Setting-2 adopts $(\alpha,\beta)=(0.2,1.0)$, whilst Setting-1 takes $(\alpha,\beta)=(0.2,1.5)$ in {ABD-MRI} and $(\alpha,\beta)=(0.3,1.0)$ in {ABD-CT}. 
For the {CMR} dataset, Setting-1 selects $(\alpha,\beta)=(0.2,1.5)$. 
As for $N$, on the three datasets, Setting-1 and Setting-2 set 8 and 16, respectively.

\paragraph{Evaluation protocol} 
To evaluate the performance of the segmentation model, we utilized the conventional Dice score scheme. The Dice score has a range from 0 to 100, where 0 represents a complete mismatch between the prediction and ground truth, while 100 signifies a perfect match. The Dice calculation formula is\\
\[ \text{Dice}(A, B) = \frac{2 \left||A \cap B\right||}{\left||A\right|| + \left||B\right||} \times 100\% \]\\
where A and B are the predicted mask and ground truth, respectively.

\begin{figure}[!h]
    \centering
    \includegraphics[width=0.45\linewidth]{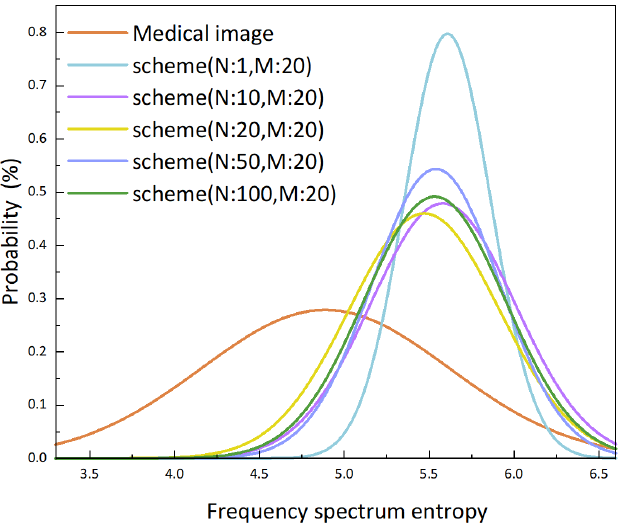}~
    \includegraphics[width=0.45\linewidth]{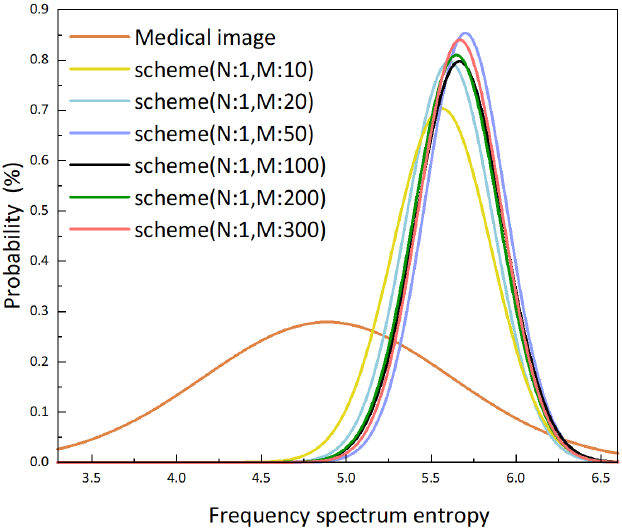}
    \caption{Comparison results of probability distribution of frequency spectrum entropy where ``scheme" is a setting to build natural image group. 
    {\bf Top:} Results as category quantity varying; 
    {\bf Bottom:} Results as image quantity varying in the same category.} 
    \label{fig:fpe}
\end{figure}

\begin{figure*}[!h]
    \centering 
    \includegraphics[width=0.98\linewidth]{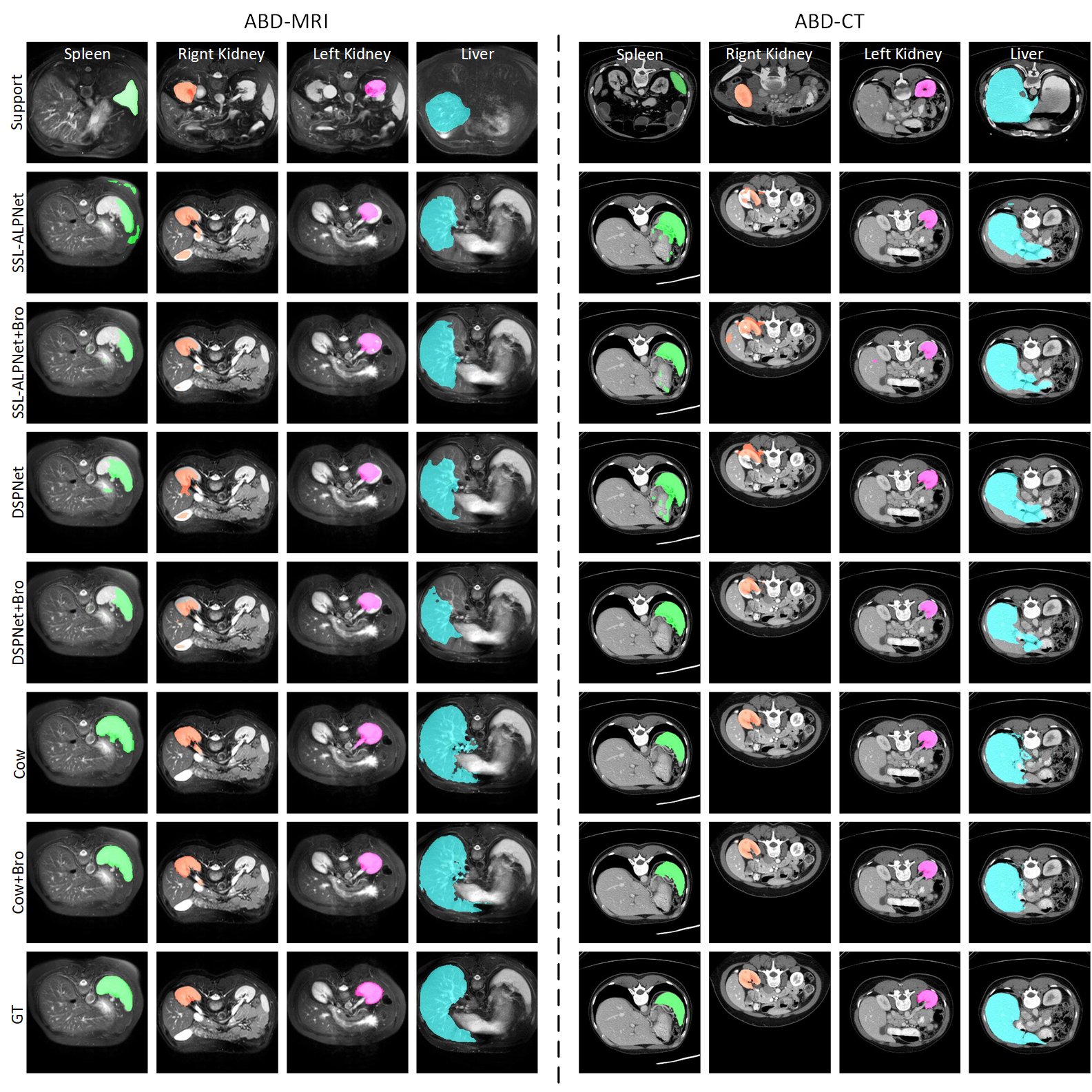}
    \caption{The qualitative comparison results on {\bf ABD-MRI} (the left side) and {\bf ABD-CT} (the right side) under Setting-2. {\bf Top} to {\bf bottom}: Support images, segmentation results and ground-truth segmentation of a query slice containing the target object (Best viewed with zoom).} 
    \label{fig:vis-abd}
\end{figure*}

\begin{figure*}[!htbp]
    \centering 
    \includegraphics[width=0.95\linewidth, keepaspectratio]{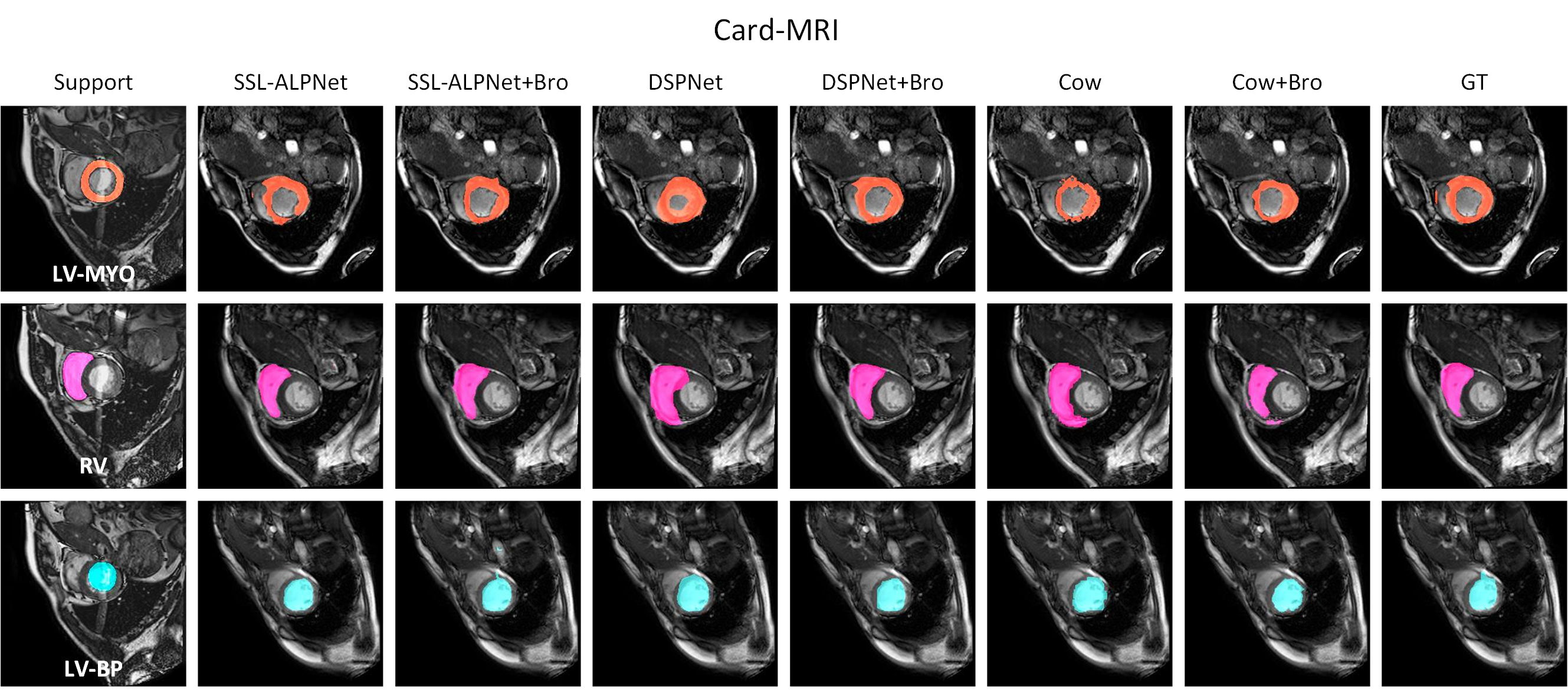} 
    \caption{
        The qualitative comparison results in {\bf CMR} under Setting-1. 
        {\bf Left} to {\bf right}: Support images, segmentation results and ground-truth segmentation of a query slice containing the target object. 
        {\bf Top} to {\bf bottom}: LV-MYO (left ventricular myocardium), RV (right ventricle) and LV-BP (left ventricular outflow tract blood pool). (Best viewed with zoom) 
        }
    \label{fig:cmr}
\end{figure*}

\subsection{Empirical Verification of Medical and Natural Images in Probability Distribution of Frequency Spectrum Entropy} \label{sec:pdfse}
\paragraph{Data preparation} 
Our experiment data includes the natural image group and the medical image group.
The natural image group consists of $N$ categories randomly selected from the ImageNet dataset~\citep{5206848}, and each category contains $M$ images taken randomly. 
In this way, we have $N*M$ natural images. 
Similarly, the medical image group consists of all 100 images from the {\bf ABD-CT}, {\bf ABD-MRI} and {\bf CMR} datasets introduced above.

\paragraph{Probability distribution of frequency spectrum entropy}
This distribution is created in three steps, applied to a group of images. First, we convert the images in this group to grayscale images and then calculate their magnitude spectrum across various frequencies. 
Following this, treating the spectrum as a distribution, we compute the corresponding information entropy value, which we refer to as frequency spectrum entropy, for each image. 
As a result, we obtain ${N*M}$ or {100} values of frequency spectrum entropy.       
In the end, we estimate the probability density function (PDF) using the in-built Matlab function ``${\rm{normpdf}}(\cdot)$"~\citep{mathworks_normpdf}.

\paragraph{Comparison results}
To exclude the impact of category, we build the natural image group using five schemes: $(N=1, M=20)$, $(N=10, M=20)$, $(N=20, M=20)$, $(N=50, M=20)$, and $(N=100, M=20)$, respectively. 
On the other hand, to exclude the impact of image quantity, we also provide the probability distribution of frequency spectrum entropy as $(N=1, M=10)$, $(N=1, M=20)$, $(N=1, M=50)$, $(N=1, M=100)$, $(N=1, M=200)$ and $(N=1, M=300)$, respectively. 

As shown in Fig.~\ref{fig:fpe}, natural images have a higher mean value but with a smaller variance, while medical images appear in a reverse situation, regardless of changes in categories (the top sub-figure) or quantities (the bottom sub-figure).
The results indicate that {\it natural images are significantly and robustly different from medical images, providing empirical evidence for our design highlighting the detailed background representation.}



\subsection{Quantitative and Qualitative Results}
\paragraph{Competitors}
To evaluate the proposed method, we choose seven state-of-the-art methods for medical image semantic segmentation as comparisons, 
including 
PANet~\citep{wang2019panet}, 
SSL-ALPNet~\citep{ouyang2022self},
ADNet~\citep{hansen2022anomaly},
RPTNet~\citep{zhu2023rpt},
Q-Net~\citep{shen2023qnet},
CAT-Net~\citep{lin2023few},
GMRD~\citep{cheng2024few},
DSPNet~\citep{Tang2024FewShotMI},
and COW~\citep{jiang2025cow}. 
All of them are prototypic approaches.

As we stated earlier, our segmentation model is formed by plugging {\shortmodelname} into SSL-ALPNet. 
Thus, we specify it as "SSL-ALPNet+{\shortmodelname}".
Also, we plug {\shortmodelname} with another model DSPNet and COW, termed DSPNet+{\shortmodelname} and COW+{\shortmodelname}. 
For a fair comparison, we obtain their results by re-running their official codes on the same evaluation bed as SSL-ALPNet+{\shortmodelname}. 
For our experimental results, we use the stochastic gradient descent algorithm with a batch size of 1 for 100k iterations to minimize the objective formulated in Eq.~\eqref{eqn:loss_final}.  
The self-supervised training is executed with Pytorch coding on a single GPU of NVIDIA 4080 took approximately 1.4 hours and consumed around 6.1GB of memory.

\paragraph{Comparison results}
Tab.~\ref{tab:rlt-ABD}$\sim$Tab.~\ref{tab:rlt-CMR} present the quantitative results for three evaluation datasets. In Setting-1, the mean accuracy of SSL-ALPNet+{\shortmodelname} surpasses that of SSL-ALPNet alone by {\bf 1.6}\%, {\bf 3.4}\%, and {\bf 1.0}\% on the ABD-MRI, ABD-CT, and CMR datasets, respectively. In Setting-2, SSL-ALPNet+{\shortmodelname} improves mean accuracy by {\bf 5.2}\% and {\bf 1.7}\% compared to SSL-ALPNet on the ABD-MRI and ABD-CT datasets, respectively. Notably, SSL-ALPNet+{\shortmodelname} demonstrates significant improvement in challenging categories, such as the Right and Left kidneys. For example, in Setting-2, SSL-ALPNet+{\shortmodelname} achieves over {80}\% accuracy on the ABD-MRI dataset. 
Similar to the SSL-ALPNet group, DSPNet+{\shortmodelname} and COW+{\shortmodelname} outperform DSPNet and COW, respectively, on all datasets in mean dice score. 
These results indicate that {\shortmodelname} effectively enhances previous approaches, as our fusion strategy provides a stronger representation of the background. 
Furthermore, SSL-ALPNet+{\shortmodelname} and DSPNet+{\shortmodelname} show a competitive advantage over other models.

For an intuitive observation, we present qualitative results in Fig.~\ref{fig:vis-abd} and Fig.~\ref{fig:cmr}. The segmentation results from SSL-ALPNet/DSPNet often include unintended background regions, such as the Right kidney in ABD-MRI (shown on the left side of Fig.~\ref{fig:vis-abd}), the Spleen in ABD-CT (on the right side of Fig.~\ref{fig:vis-abd}), and the LV-BP in CMR (Fig.~\ref{fig:cmr}). In contrast, SSL-ALPNet+{\shortmodelname} and DSPNet+{\shortmodelname} significantly reduce these segmentation errors. This comparison confirms that the proposed background-fused prototypes can improve the distinction between foreground and background.

\begin{figure}[!h]
    \setlength{\belowcaptionskip}{0pt}
    \setlength{\abovecaptionskip}{0pt}
    \begin{center}
        \includegraphics[width=0.98\linewidth]{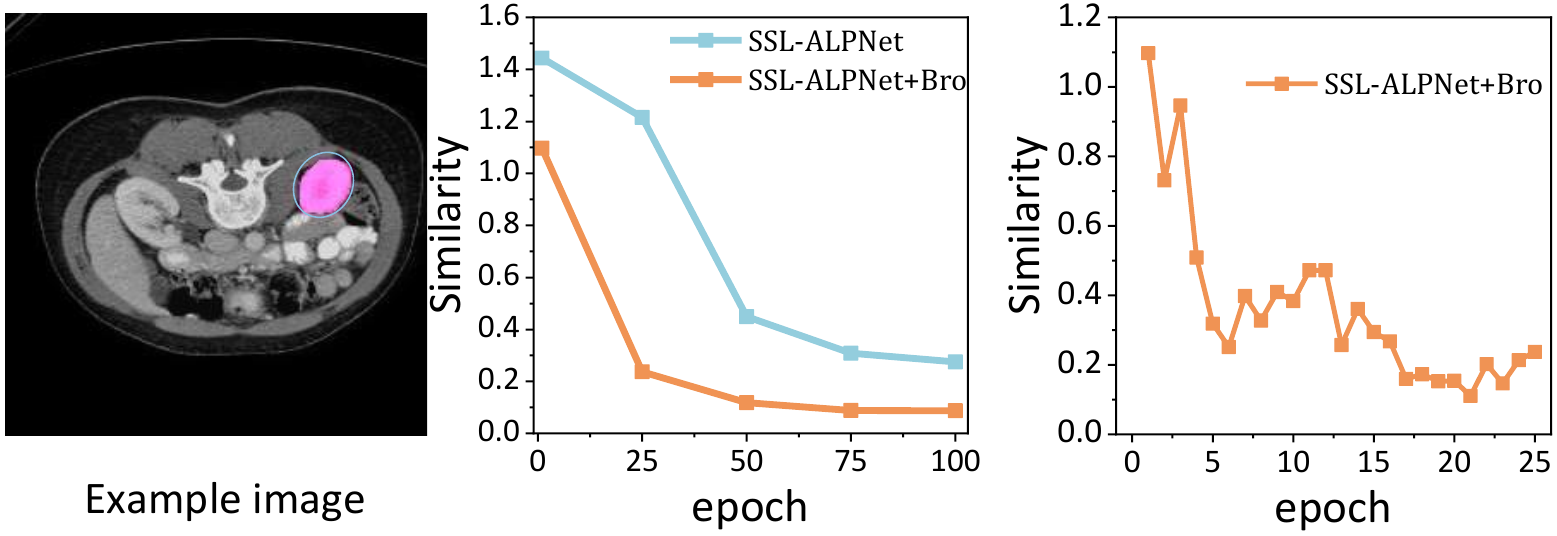}
    \end{center}
    \caption{ 
    Background representation analysis on {\bf ABD-CT} in Setting-1. 
    {\bf Left:} Example image with foreground marked in purple; {\bf Middle:} Variation of convolutional similarity between the background-fused prototypes and the foreground over epoch 3$\sim$100 with gap 25; {\bf Right:} Details as epoch varies from 3 to 25. 
    }
    \label{fig:bg-confu}
\end{figure}

\begin{table*}[!t]
    \caption{Ablation study results on {\bf ABD-CT}.
    Here, Bro is short for SSL-ALPNet+{\shortmodelname}.
    Numbers in bold indicate the best results.} 
    \label{tab:ABLA}
    \renewcommand\tabcolsep{8pt} 
    \renewcommand\arraystretch{1.2}
    \centering
    \scriptsize
    \begin{adjustbox}{max width=\textwidth}
    \begin{tabular}{l|l|c c| c c c c c | c c c c c}
        \toprule
        \multirow{2}{*}{\textbf{\#}} &\multirow{2}{*}{\textbf{Method}} &\multirow{2}{*}{${\rm{B}}_{\triangle}$} &\multirow{2}{*}{${\mathcal{L}_{adv}}$} &\multicolumn{5}{c|}{\textbf{Setting-1}}  &\multicolumn{5}{c}{\textbf{Setting-2}}\\
        
         & & & &{Liver} &{R.kidney} &{L.kidney} &{Spleen} &{Mean} &{Liver} &{R.kidney} &{L.kidney} &{Spleen} &{Mean}  \\
         \midrule
        1 & SSL-ALPNet & -- & -- & 67.29 & 72.62 & 76.35 & 70.11 & 71.59 & 69.14 & 59.05 & 64.18 & 61.97 & 63.59 \\
        \midrule
        2 & {\shortmodelname} w/o FeaC & -- & -- &70.29  &75.98  &78.21  & 69.77 &73.56  &59.95  & 55.63 &72.50  &64.12  &63.05  \\
        3 & {\shortmodelname} w/o HiCA & -- & -- &67.93 &77.27  &79.58  &68.88  &73.48  &57.78  &61.83  &69.70  &64.58  &63.47  \\
        \midrule
        4 & {\shortmodelname} w/o AD   & \xmark & \xmark & 69.71 & 75.94 & 73.39 & 72.41 & 72.86 & 54.92 & 57.39 & 61.38 & 68.72 & 60.60 \\
        5 & {\shortmodelname} w/o AD-$B_{\triangle}$ & \xmark & \cmark & 61.70 & 71.52 & 74.19 & 70.50 & 69.48 & 54.83 & 57.9 & 65.11 & 66.47 & 61.08 \\
        6 & {\shortmodelname} w/o AD-${\mathcal{L}_{adv}}$ & \cmark & \xmark & 66.22 & 72.81 & 74.62 & 69.97 & 70.90 & 54.93 & 56.27 & 64.91 & 63.16 & 59.82 \\ 
        7 & {\shortmodelname} & \cmark & \cmark & \textcolor{blue}{\bf 71.01}  &
        \textcolor{blue}{\bf 79.07} & \textcolor{blue}{\bf 77.91} & \textcolor{blue}{\bf 71.71} & \textcolor{blue}{\textbf{\textit{74.93}}} & \textcolor{blue}{\bf 60.81} & \textcolor{blue}{\bf 66.82} & \textcolor{blue}{\bf 65.49} & \textcolor{blue}{\bf 67.87} & \textcolor{blue}{\textbf{65.25}} \\        
        \bottomrule
    \end{tabular}
    \end{adjustbox}
\end{table*}

\subsection{Analysis of Background Representation} 
This section presents an empirical analysis of the representation of the background, using an example image shown on the left side of Fig.~\ref{fig:bg-confu}.
During the model training with 100 epochs, we set checkpoints at epochs 3, 25, 50, 75, and 100, obtaining five intermediate models. Inputting the example image into these intermediate models, we get corresponding five pairs of image's feature maps and background prototypes, denoted $\{F_k, P_k\}_{k=1}^{5}$. 
Suppose that the right kidney in this example image (masking in purple) is in the foreground. 
We compute the similarity between $P_k$ and the foreground zone in $F_k$ based on convolution computing, leading to a similarity variation curve, as shown in the middle of Fig.~\ref{fig:bg-confu}. Meanwhile, we choose the SSL-ALPNet method as a comparison.

It is seen that SSL-ALPNet has a similarity decline due to the work of foreground prototype, and the introduction of {\shortmodelname}-based background representation brings out more noticeable decreases. 
For a clear view, we also provide the variation from epoch 3 to 25 on the right side of Fig.~\ref{fig:bg-confu}. 
Those results suggest that {\shortmodelname} leads to a detailed representation of background as expected.
In addition, {\shortmodelname} essentially encourages a global fusion. 



\begin{figure}[!htbp]
    \centering 
    \includegraphics[width=0.8\linewidth]{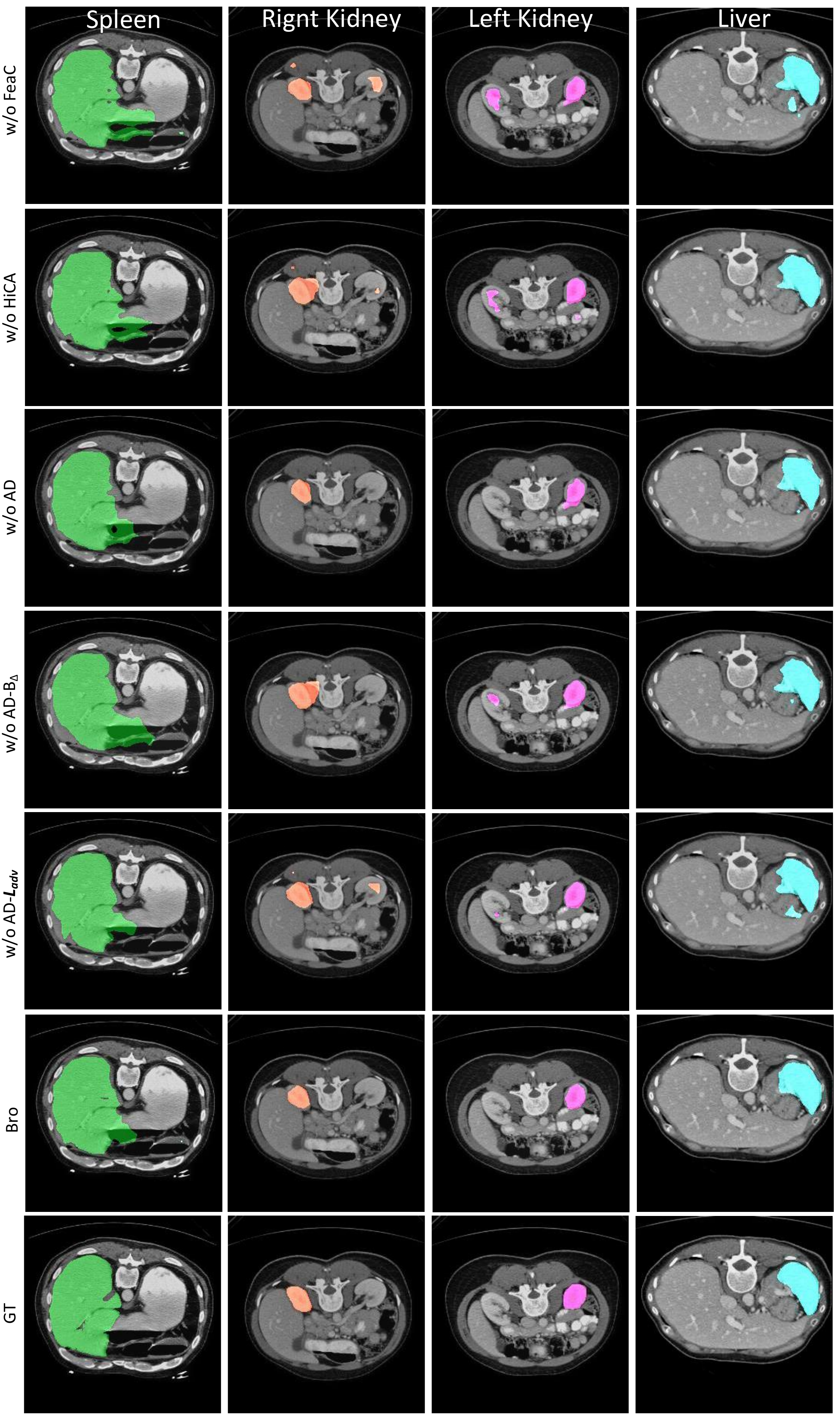}
    \caption{The qualitative comparison results of ablation study in the {\bf ABD-CT} dataset under Setting-2, in which Bro is short for SSL-ALPNet+{\shortmodelname}. 
    {\bf Left} to {\bf right}: Liver, Right kidney, Left kidney, and Spleen. (Best viewed with zoom)} 
    \label{fig:ablation}
\end{figure}

\subsection{Further Model Analysis} 
In this part, for a clear view, all analysis experiments are conducted based on the SSL-ALPNet+{\shortmodelname} method.

\paragraph{Ablation study} 
This part isolates the effect of (1) FeaC, (2) HiCA and (3) adversarial regularization (AD) in HiCA. Our experiments take SSL-ALPNet as baseline.  
First, by removing FeaC and HiCA from SSL-ALPNet+{\shortmodelname}, respectively, we buid two variarions SSL-ALPNet+{\shortmodelname} w/o FeaC and SSL-ALPNet+{\shortmodelname} w/o HiCA.  
Compared with SSL-ALPNet+{\shortmodelname}, the two variations decrease on mean accuracy by {\bf 1.5}\% at least in Setting-1 and Setting-2, confirming the effect of FeaC and HiCA. 
Moreover, as FeaC or HiCA works alone, the performance is close to SSL-ALPNet, indicating that FeaC and HiCA reinforce each other and jointly contributes to the final performance.  

To evaluate the AD design, we remove it from HiCA and refer to the modified method as SSL-ALPNet+{\shortmodelname} w/o AD where the weakened HiCA degenerates to self-cross attention. 
This removal leads to a decline of {\bf 2.1}\% in Setting-1 and {\bf 4.6}\% in Setting-2 from SSL-ALPNet+{\shortmodelname}, highlighting the significance of AD.

To better understand this design, we further elaborate the effect of AD's components, i.e., $B_{\triangle}$ and $\mathcal{L}_{adv}$. Correspondingly, by removing them, respectively, we have two variation methods SSL-ALPNet+{\shortmodelname} w/o AD-$B_{\triangle}$ and SSL-ALPNet+{\shortmodelname} w/o AD-$\mathcal{L}_{adv}$. 
Their effect is verified by the evident performance decrease. 
In particular, when only $B_{\triangle}$ is available, SSL-ALPNet+{\shortmodelname} w/o AD-$\mathcal{L}_{adv}$ have a mean accuracy decline of {\bf 4.0}\% at least on Setting-1 and Setting-2, compared with SSL-ALPNet+{\shortmodelname}, even behind SSL-ALPNet. 
These results show that single usage of $B_{\triangle}$ might make adjustments out of control, also providing empirical evidence for the necessity of $B_{\triangle}$. 
Meanwhile, using $\mathcal{L}_{adv}$ alone plays a negative role. 
For example, SSL-ALPNet+{\shortmodelname} w/o AD-$B_{\triangle}$'s result ({\bf 69.48}\% in Setting-1, {\bf 61.08}\% in Setting-2) is worse than SSL-ALPNet ({\bf 71.59}\% in Setting-1, {\bf 63.59}\% in Setting-2). 
The findings indicate that $B_{\triangle}$, $\mathcal{L}_{adv}$ only make sense in the adversarial context.

As a supplement to the ablation study above, we present qualitative results in Fig.~\ref{fig:ablation}. 
There are three observations. 
First, when FeaC or HiCA operate independently (as seen in the first and second rows), the segmentation results frequently include regions that deviate from the correct segmentation areas. This highlights the contributions of denoising (from FeaC) and the background fusion strategy (from HiCA). Second, when we remove the adversarial structure from HiCA (illustrated in the third row), the results no longer include far-located regions. This demonstrates the effectiveness of the coarse-grained attention mechanism based on channel groups. However, the segmentation boundaries still diverge substantially from the ground truth, indicating the importance of the fine-grained adjustments provided by the adversarial structure. Third, when either $B_{\triangle}$ or $L_{adv}$ is used independently (shown in the fourth and fifth rows), the results are poorer compared to the scenario without adversarial regularization (w/o AD). 
In contrast, when $B_{\triangle}$ and $L_{adv}$ are combined to form the adversarial regularization, i.e., SSL-ALPNet+Bro (as depicted in the second-to-last row), the segmentation performance is significantly improved and reaches its best level.
\begin{figure}[!h]
    \begin{center}
        \includegraphics[width=0.32\linewidth]{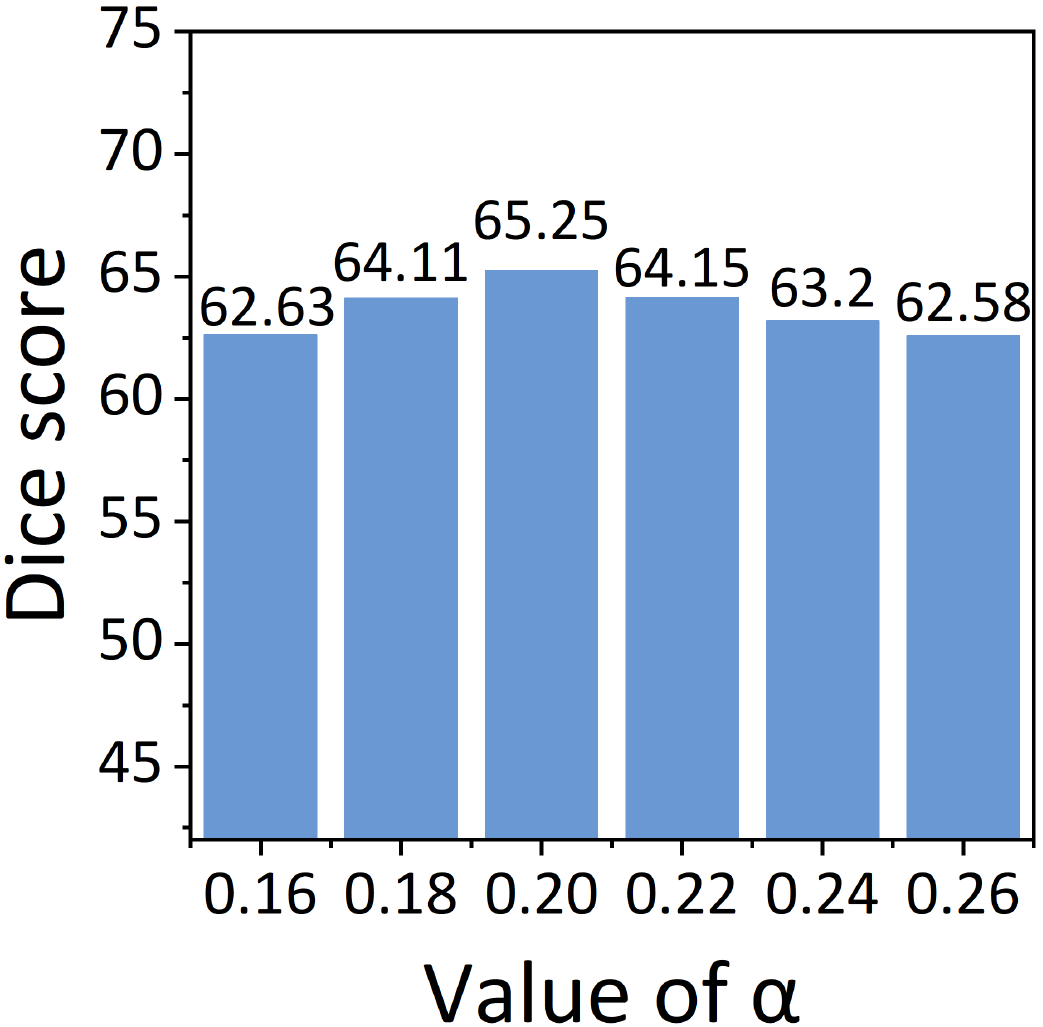}
        \includegraphics[width=0.32\linewidth]{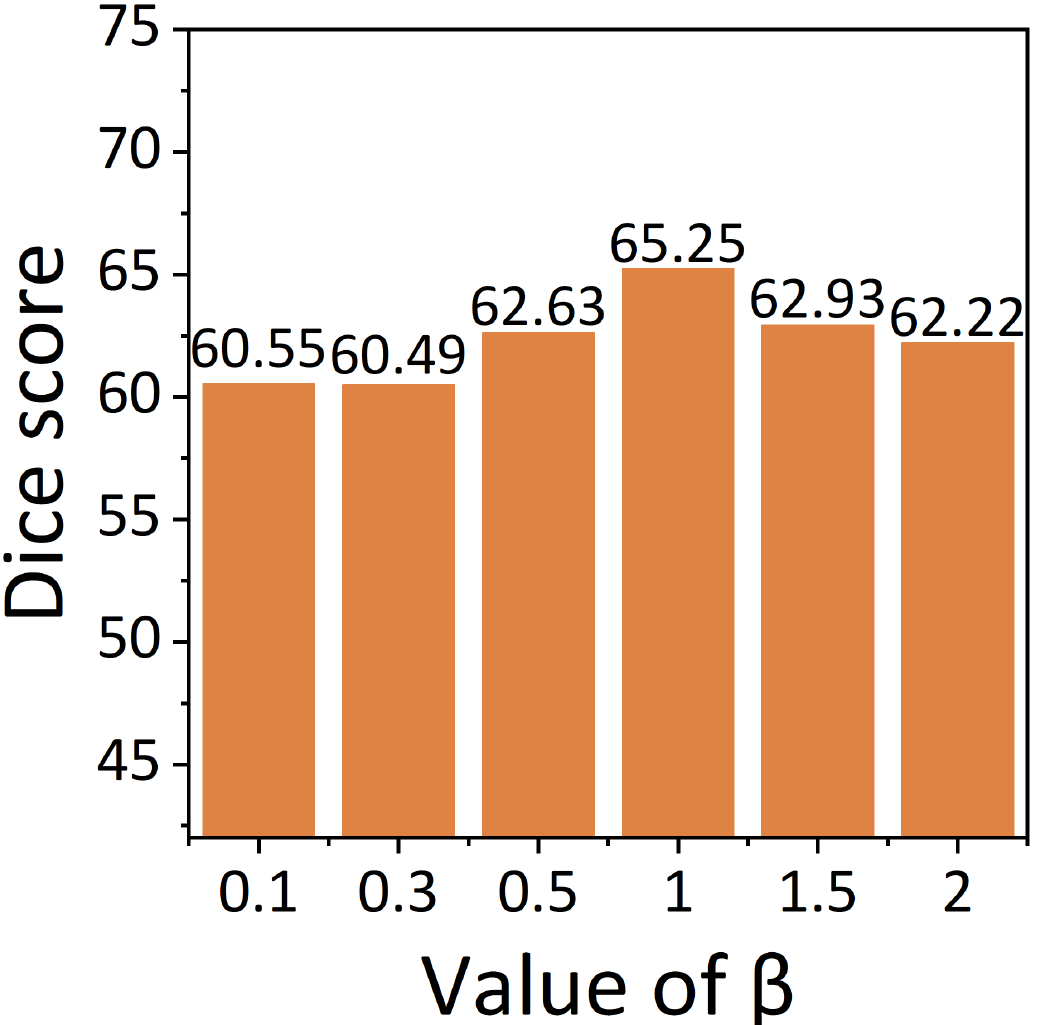}
        \includegraphics[width=0.32\linewidth]{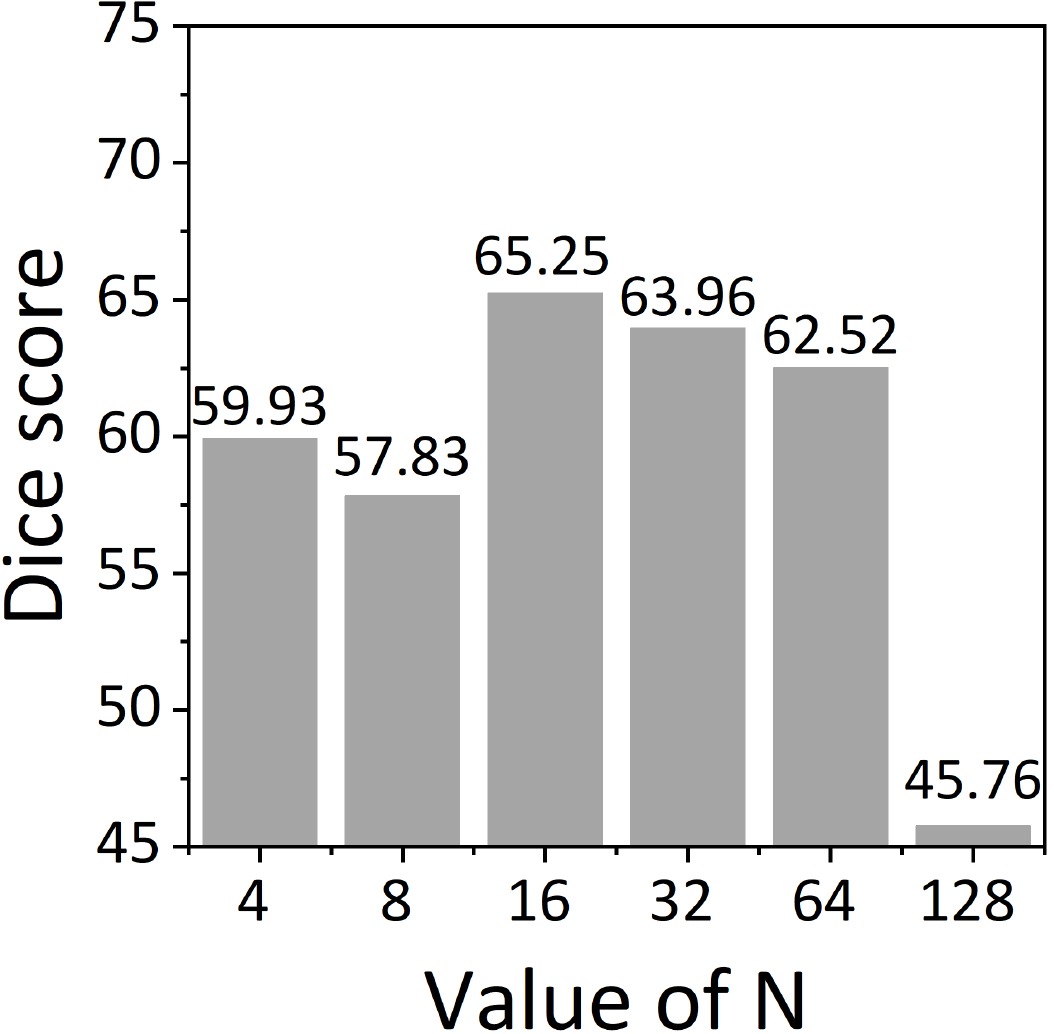}
    \end{center}
    \caption{ 
    Performance variation as parameter varying on the {\bf ABD-CT} dataset in Setting-2. {\bf Left, Middle} and {\bf Right} are results of $\alpha$, $\beta$ and $N$, respectively. 
    }
    \label{fig:param-sen}
\end{figure}


\paragraph{Parameter sensitiveness.} 
This section examines the impact of three parameters in {\shortmodelname}: $\alpha$ in Eq.~\eqref{eqn:hica}, $\beta$ in Eq.~\eqref{eqn:loss_final}, and the grouping parameter $N$.  Fig.~\ref{fig:param-sen} illustrates how the performance, measured by the Dice score, changes as these parameters vary. As indicated on the left and in the middle of the figure, the performance does not experience significant vibration, suggesting that it is relatively insensitive to the values of $\alpha$ and $\beta$.  However, on the right side of the figure, it is evident that when $N$ is either too small or too large, the performance declines noticeably. 
For instance, the Dice score drops by {\bf 19.5}\% at $N=128$ compared to the score at $N=16$. This decline can be explained by the fact that both small and large grouping sizes disrupt the representational connection between the channel groups and their corresponding semantics.


\vspace{-2.8mm}

\section{Conclusion} 

Unlike the clear separation between foreground and background in natural images, medical images often exhibit similar visual features in both foreground and background (reflected in a concentrated frequency distribution), making distinction difficult. This paper introduces a novel pluggable approach for FSS in medical images, referred to as {\shortmodelname}.
In the proposed pipeline, the FeaC and HiCA modules jointly contribute to the background fusion in the support image. 
After FeaC filters out noises, HiCA refines the background-fused prototypes by a coarse-to-fine attention mechanism over different channel groups. 
To accomplish this, we propose a trainable Mean-Offset structure with adversarial regularization. 
{\shortmodelname}’s effectiveness is validated by SOTA results across three challenging medical datasets.

\section*{Declaration of competing interest} 
The authors declare that they have no known competing financial interests or personal relationships that could have appeared to influence the work reported in this paper.

\section*{Acknowledgments}
\begin{sloppypar}
\small
Supported by National Key R\&D Program of China (2022YFB4700700), the National Natural Science Foundation of China (U24A20672), the Yunnan Provincial Department of Science and Technology Social Development Special Project (202403AC100003), and CAMS Innovation Fund for Medical Sciences (CIFMS), 2025-I2M-C\&T-A-003.
\end{sloppypar}

\FloatBarrier



\bibliographystyle{elsarticle-num}
\bibliography{cas-refs-comm}




\end{document}